# Revolutionizing Process Mining: A Novel Architecture for ChatGPT Integration and Enhanced User Experience through Optimized Prompt Engineering

Mehrdad Agha Mohammad Ali Kermani [a,*]

Hamid Reza Seddighi [a]

Mehrdad Maghsoudi [b]

[b] *Department of Management, Economics and Progress Engineering, Iran University of Science and Technology, Iran, Tehran, Iran*
[a] *Department of Industrial and Information Management, Faculty of Management and Accounting, Shahid Beheshti University, Tehran, Iran*

**Abstract**

In the rapidly evolving field of business process management, there is a growing need for analytical tools that can transform complex data into actionable insights. This research introduces a novel approach by integrating Large Language Models (LLMs), such as ChatGPT, into process mining tools, making process analytics more accessible to a wider audience. The study aims to investigate how ChatGPT enhances analytical capabilities, improves user experience, increases accessibility, and optimizes the architectural frameworks of process mining tools.

The key innovation of this research lies in developing a tailored prompt engineering strategy for each process mining submodule, ensuring that the AI-generated outputs are accurate and relevant to the context. The integration architecture follows an Extract, Transform, Load (ETL) process, which includes various process mining engine modules and utilizes zero-shot and optimized prompt engineering techniques. ChatGPT is connected via APIs and receives structured outputs from the process mining modules, enabling conversational interactions.

To validate the effectiveness of this approach, the researchers used data from 17 companies that employ BehfaLab's Process Mining Tool. The results showed significant improvements in user experience, with an expert panel rating 72% of the results as "Good". This research contributes to the advancement of business process analysis methodologies by combining process mining with artificial intelligence.

Future research directions include further optimization of prompt engineering, exploration of integration with other AI technologies, and assessment of scalability across various business environments. This study paves the way for continuous innovation at the intersection of process mining and artificial intelligence, promising to revolutionize the way businesses analyze and optimize their processes.

---

[*] Corresponding author. e-mail: m_kermani@iust.ac.ir

## 1. Introduction

In the dynamic realm of business process management, the digital revolution has ushered in an urgent demand for analytical tools capable of transforming intricate data landscapes into actionable business insights. Process Mining (PM) emerges as a pivotal innovation in this context, offering a transparent and comprehensive perspective on operational efficiencies and bottlenecks within business process event logs (Van Der Aalst, 2016; Van der Aalst et al., 2012). However, the full potential of PM often remains underutilized, hindered by the necessity for specialized expertise to interpret its complex outputs.

This research is positioned at the heart of this challenge, aiming to democratize the power of process mining through the integration of Large Language Models (LLMs) like ChatGPT. These models hold the promise of bridging the gap between the intricate domain of process analytics and the diverse user base that can benefit from such insights (Buijs et al., 2014; Dumas et al., 2018). The integration of LLMs into process mining tools is a pioneering step, poised to transform intricate data analysis into intuitive conversational interactions, accessible even to non-experts.

The study is guided by the following research questions:
1. How do LLMs, particularly ChatGPT, augment the analytical capabilities of traditional process mining tools?
2. What impact does the integration of conversational AI have on the user experience in process mining?
3. How does the incorporation of LLMs enhance the accessibility and usability of process mining tools for a wider audience?
4. What architectural framework is optimal for integrating LLMs with process mining tools to maximize their effectiveness?
5. What is the effect of prompt engineering on the quality of outputs?

This research marks a significant departure from traditional process mining methodologies by innovatively applying ChatGPT, an LLM, to this field. This groundbreaking integration not only deviates from established practices but also sets a new precedent in the utilization of conversational AI for business analytics. The novelty of this approach is underscored by its alignment with the evolving trends in process mining and AI, as discussed in foundational works (Mannhardt et al., 2016; Van Der Aalst, 2012).

A key innovation in this study is the development of a specialized prompt engineering strategy, meticulously tailored to each process mining submodule. This strategy ensures that AI-generated outputs are not only accurate but also contextually relevant, addressing the nuanced requirements of process mining tasks. This approach draws upon the principles established in the prior works of (van Dongen et al., 2005; Weijters et al., 2006), which emphasize the importance of detailed and context-specific analysis in process mining.

The research adopts a detailed, submodule-focused implementation methodology, facilitating seamless integration of ChatGPT into the existing process mining framework. This methodological approach enhances each component's functionality while maintaining the system's overall coherence, resonating with the insights provided earlier (Bose et al., 2013; Van Der Aalst, 2012). Additionally, the study presents an architectural blueprint for integrating LLMs within process mining tools, potentially serving as a model for future systems.

The practical implications of this research are substantial, contributing to enhanced user experience, heightened customer satisfaction, and more informed decision-making in businesses. The empirical validation of the AI-enhanced PM tool, using data from 17 companies, underscores the study's relevance and potential for broad application, aligning with the prior findings. The study also anticipates future developments in prompt engineering, as AI technology continues to advance, drawing upon the forward-looking perspectives (Van Der Aalst, 2016; Van der Aalst et al., 2012).

The subsequent sections of the study will detail the methodology behind integrating ChatGPT with process mining tools, present empirical findings from various business contexts, and discuss the broader implications for the field of process mining. The conclusion will synthesize these insights and explore future research directions, envisioning a future where AI is an indispensable and transformative element in process mining.

## 2. Literature Review
### 2.1. Process Mining Lifecycle

Process mining is a discipline that sits at the intersection of data science and process science (Van Der Aalst & van der Aalst, 2016). Its goal is to extract knowledge and insights from event data recorded by information systems in order to understand, analyze, and improve real-world business processes (van der Aalst & Carmona, 2022).

At a high level, the process mining lifecycle consists of the following main steps, as depicted in Figure 1:

1. Extract event data from information systems
2. Explore, select, filter, and clean the event data
3. Discover process models from the event data
4. Check conformance between modeled and observed behavior
5. Enrich process models with insights around performance, predictions, improvements
6. Transform insights into actions through modeling, adaptation, and showing results

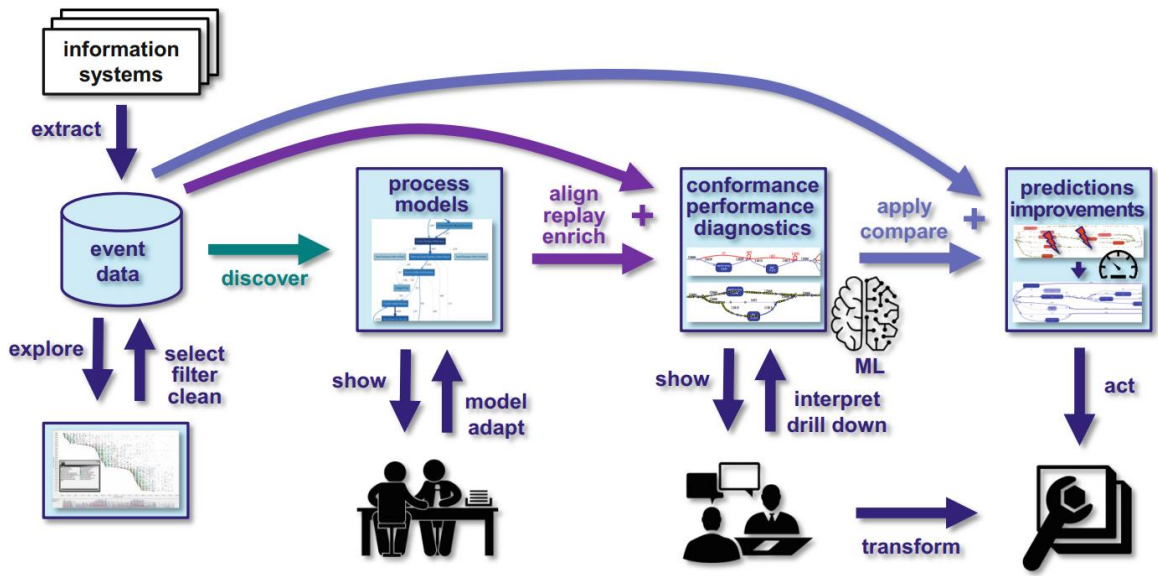

*Figure 1: Process Mining Lifecycle (van der Aalst & Carmona, 2022)*

The starting point is extracting event data from the various operational systems that support a business process. Each event refers to a specific activity occurring for a particular process instance (case) at a certain time. Finding, extracting, and transforming this raw data into a standardized event log format suitable for process mining is still a significant challenge in practice, often taking up to 80% of the effort in a process mining project.

Once an event log is available, the next step is to explore the data to understand its characteristics and quality. This involves filtering the data to focus on the relevant scope, handling missing or duplicate events, and dealing with complex event logs involving intertwined processes and diverse case notions. More advanced techniques like object-centric process mining can help with the latter.

The core process mining techniques can then be applied, starting with automated process discovery. The goal is to learn a process model from example behavior seen in the event log, capturing the main process flows and decision points. While powerful techniques exist, process discovery is not a completely solved problem, especially when it comes to dealing with concurrency and constructing models that are accurate yet also simple and understandable. Combining algorithmic approaches with interactive user guidance is an important direction for future work.

Next, conformance-checking techniques can be used to analyze discrepancies between modeled and observed behavior, highlighting deviations at the case level. Efficient conformance checking for large and complex models and event logs remains computationally challenging.

Process mining diagnostics can also provide insights beyond conformance, for example, identifying bottlenecks and inefficiencies. Performance metrics like flow time, waiting time, and service time can be extracted and visualized. Applying machine learning to event data annotated with such information allows for predicting future behavior and suggesting process enhancements, although this requires considering a richer feature set beyond just the activity sequences.

Ultimately, the goal is to turn these process-mining insights into concrete improvement actions. On the one hand, this requires presenting the results in an understandable way to process stakeholders through interactive visualization. On the other hand, process mining tools are increasingly integrating with automation platforms to automatically trigger actions such as launching remedial workflows to handle problematic cases. Explaining and recommending process model changes is another area where academic research can have a real-world impact (van der Aalst & Carmona, 2022).

### 2.2. Process Mining and Artificial Intelligence

The integration of Artificial Intelligence with Process Mining marks a significant milestone in the realm of process analysis. This fusion heralds a new era by endowing PM tools with advanced predictive analysis and optimization capabilities, rendering them more adaptive and intelligent (Chapela-Campa & Dumas, 2023; Mehdiyev & Fettke, 2021a). According to (Dumas et al., 2018), this collaboration empowers organizations to anticipate future trends and behaviors within their processes (Mehdiyev & Fettke, 2021a). The role of AI in Process Mining extends beyond mere analysis to the realm of automated discovery and continual improvement of process models (Chapela-Campa & Dumas, 2023). Notably, machine learning algorithms play a pivotal role in this synergy, enabling the extraction of intricate, non-linear process patterns (Folino & Pontieri, 2021). This capability provides organizations with deeper insights and more accurate predictions, as highlighted by (Van Der Aalst, 2016).

The marriage of AI and PM creates a dynamic synergy that not only enhances current process analysis but also propels organizations into the realm of proactive decision-making (Warmuth & Leopold, 2022). The predictive capabilities infused by AI enable organizations to not only understand historical process data but also anticipate and adapt to future changes seamlessly (Mehdiyev & Fettke, 2021b). This paradigm shift in process analysis equips organizations with a strategic advantage, allowing them to stay ahead of the curve and optimize their operations effectively(Mehdiyev & Fettke, 2021b).

Moreover, the integration of AI in Process Mining has been demonstrated to enhance anomaly detection and root cause analysis. By leveraging advanced algorithms, organizations can identify deviations from expected process patterns in real time, facilitating swift intervention and resolution of issues. This proactive approach to problem-solving minimizes operational disruptions and contributes to overall efficiency (Folino & Pontieri, 2021; Mehdiyev & Fettke, 2021b).

Additionally, the utilization of Natural Language Processing (NLP) in combination with Process Mining enhances the understanding of unstructured data, such as textual information within documents and communication logs. This not only enriches the analysis but also opens avenues for a more comprehensive understanding of the underlying processes (Barbieri et al., 2023; Yeo et al., 2022).

Furthermore, the continuous learning aspect of AI in Process Mining allows for the adaptation of models to evolving processes. As new data is generated and patterns change, AI algorithms can dynamically adjust, ensuring that process models remain relevant and effective over time (Chaima & KHEBIZI, 2022; Pery et al., 2021). This adaptability contributes to the sustainability of the process optimization efforts undertaken by organizations.

### 2.3. Generative AI: Overview and Impact

Generative AI refers to machine learning systems that can generate new content and artifacts such as text, images, audio, and video. Unlike traditional AI systems that are trained to classify inputs or predict outputs, generative models like GPT-3 and DALL-E can synthesize high-quality outputs from simple text prompts without direct instruction. This emergent capability has exciting potential across many domains but also poses risks if deployed irresponsibly (Bandi et al., 2023; Ooi et al., 2023).

Generative models are often built on deep learning architectures like transformers that are trained on massive datasets. For text generation, models like GPT-3 ingest vast corpora of natural language data to learn the statistical patterns and dependencies that allow coherent language production (Bandi et al., 2023). Image generation models like DALL-E are trained to generate plausible images from text captions by learning visual concepts from image datasets (Ooi et al., 2023). The unprecedented scale of data and computing enables these models to absorb immense world knowledge and generate human-like artifacts (Bandi et al., 2023).

Applications of generative AI include content creation, conversational agents, drug discovery (Tang et al., 2021), marketing (Kshetri et al., 2023), and code generation (Fui-Hoon Nah et al., 2023; Ooi et al., 2023). Text generation models can produce natural language for customer service bots, research paper drafts, and more. Image generation aids artists and designers by synthesizing illustrations from prompts (Fui-Hoon Nah et al., 2023). However, risks include generating misleading, biased, or toxic content. Regulations and responsible governance are needed to ensure generative AI benefits society (Fui-Hoon Nah et al., 2023; Ooi et al., 2023).

Generative AI, a subset of AI focusing on creating new content, has transformative implications across various fields. In PM, generative AI can simulate complex process scenarios, providing foresight into potential process deviations and improvements (Goodfellow et al., 2014).

In the context of PM, generative AI models offer predictive capabilities, enabling proactive process optimization. This advancement is crucial for businesses looking to adapt and thrive in dynamic market environments (LeCun et al., 2015).

Generative AI's application in engineering is vast, ranging from system design to operational process optimization. Its ability to generate novel solutions and predict outcomes enhances efficiency and innovation in engineering tasks (Goodfellow et al., 2014).

In process engineering, generative AI plays a crucial role in predictive maintenance and process control. By simulating various operational scenarios, it aids engineers in identifying optimal configurations, thus enhancing efficiency and reducing downtime (LeCun et al., 2015).

## 2.4. Integration of Generative AI with Process Mining

The rapid advancements in Artificial Intelligence (AI) and Machine Learning (ML) have opened up new avenues for enhancing business process management. Process Mining, a data-driven approach for analyzing and improving business processes, has benefited significantly from the integration of AI and ML techniques (Van Der Aalst, 2016). In particular, Generative AI has shown great promise in augmenting process mining tasks and addressing challenges related to low-quality event logs (Folino & Pontieri, 2021).

This literature review aims to provide a comprehensive overview of the recent developments in the integration of Generative AI with Process Mining. We will explore the key concepts, methodologies, and frameworks proposed in the literature, highlighting their potential benefits and challenges. Furthermore, we will discuss the role of Large Language Models (LLMs) in process mining and their implementation in the open-source library PM4Py (Berti et al., 2023a).

Generative AI, with its ability to generate new, statistically plausible outputs, has the potential to significantly enhance process mining techniques (Berti et al., 2023a; Folino & Pontieri, 2021; Kampik et al., 2023). One of the primary challenges in process mining is dealing with low-quality event logs, which can hinder the accuracy and effectiveness of the analysis. Two AI-based strategies have been proposed to address this issue: (a) using explicit domain knowledge and (b) exploiting auxiliary AI tasks [14]. These strategies can help improve the quality of process mining results, particularly when dealing with logs from low-structured processes and non-process-aware systems (Folino & Pontieri, 2021).

Large Process Models (LPMs) have emerged as a concept that combines the correlation power of Large Language Models with the analytical precision and reliability of knowledge-based systems and automated reasoning approaches (Berti et al., 2023a). LPMs leverage the vast experience of process management experts and the performance data of diverse organizations to provide a more comprehensive and accurate analysis of business processes (Berti et al., 2023a).

Several frameworks and methodologies have been proposed to facilitate the integration of Generative AI with Process Mining. Kourani et al. introduce a novel framework that utilizes LLMs to automate the generation of process models from textual descriptions. Their framework employs advanced techniques such as prompt engineering, error handling, and code generation to transform natural language process descriptions into executable models. The authors also propose an interactive feedback loop, allowing users to refine the generated models based on their input. The framework leverages the Partially Ordered Workflow Language (POWL) for intermediate process representation, ensuring the soundness of the generated models (Kampik et al., 2023).

Berti et al. investigate the application of LLMs in process mining by abstracting standard process mining artifacts and describing prompting strategies. They propose methods to create textual abstractions of event logs and process models suitable for LLMs' input constraints. The authors implement these abstraction techniques in the open-source process mining library PM4Py (Kourani et al., 2024). Through a case study using publicly available event logs and GPT-4, they assess prompts that can be directly answered by the LLM and demonstrate hypothesis formulation and verification against the entire dataset using SQL queries (Kourani et al., 2024).

Berti (Berti, 2024) examines the current paradigms for process mining on LLMs and their implementation in the PM4Py library. The author presents various techniques for abstracting process mining artifacts, such as textual and visual abstractions of event logs and process models. The paper also discusses different implementation paradigms, including direct provision of insights, translation of natural language statements to SQL queries, code generation, and automatic formulation of hypotheses (Berti, 2024).

While the integration of Generative AI with Process Mining holds great potential, it also presents certain challenges. The behavior of generative AI models can be unpredictable and at times undesirable (Berti et al., 2023a). Therefore, it is crucial to balance the correlation power of generative AI with the analytical precision of process mining techniques to ensure safety and trustworthiness (Berti et al., 2023a).

Van der Aalst emphasizes that while generative and predictive AI will significantly impact process management, dedicated process-mining approaches remain essential. Organizations need to master basic process mining techniques before advancing to more sophisticated AI and ML applications. The author identifies three areas where AI/ML techniques complement process mining: supporting the creation of event data, creating ML models to answer process-related questions, and making interactions with process mining software more human-like, easier, and supportive (van der Aalst, 2023a).

Future research should focus on addressing the challenges related to privacy, hallucinations, and the limited context window when using LLMs for process mining tasks [31]. The importance of Object-Centric Process Mining (OCPM) and Object-Centric Event Data (OCED) as key enablers for combining structured and unstructured data in a more hybrid approach should also be explored further (van der Aalst, 2023a).

## 3. Methodology

The integration of LLMs such as ChatGPT into process mining represents a significant advancement in the field, enabling more nuanced and context-aware analysis of process data. Following the approach outlined by (Berti et al., 2023b), our study adopts specific abstractions and prompt definitions tailored to the requirements of process mining. This methodology ensures that AI-generated insights are both relevant and actionable, facilitating a deeper understanding of process dynamics and potential areas for optimization. This integration is crucial for democratizing access to process mining insights, allowing users of varying technical expertise to engage in meaningful analysis and decision-making. By leveraging ChatGPT's conversational AI, PM tool not only becomes more accessible but also significantly improves in its ability to provide real-time, contextually relevant insights and recommendations. This synergy between advanced process mining techniques and cutting-edge AI ensures a more efficient, user-friendly, and insightful exploration of business processes, making it an indispensable asset for organizations aiming to optimize their operational workflows.

This section outlines an innovative architecture that integrates LLMs, specifically ChatGPT, with process mining tools to create a more intuitive and powerful tool for business process analysis. The architecture, as illustrated in the accompanying Figure 2, follows a workflow that begins with raw data extraction and culminates in delivering actionable insights through a conversational interface.

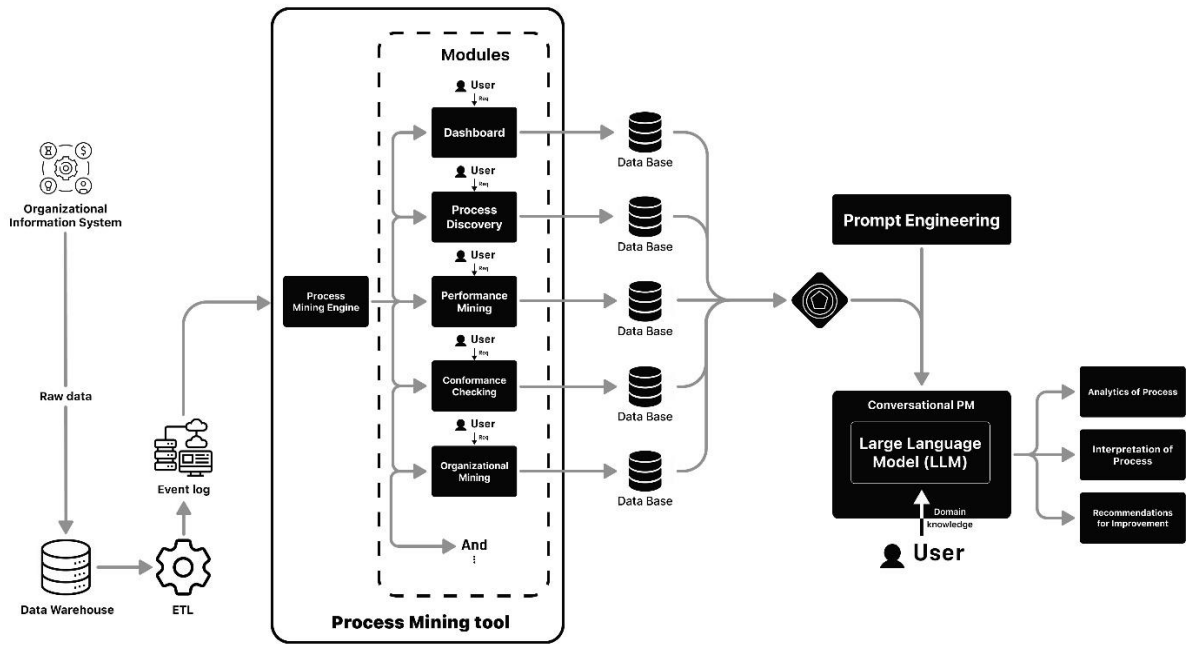

Figure 2: Proposed Architecture

**ETL and Data Warehousing**

The architecture is anchored in the ETL (Extract, Transform, Load) process. This phase involves extracting raw data from diverse information systems, transforming it to align with process mining requirements, and loading it into a data warehouse. The generated event logs are critical as they capture the detailed sequence of business process activities.

- **Process Mining Engine Modules**

Within the realm of process mining, a fundamental component lies in the Process Mining Engine, which encompasses a range of tools designed to carry out specific functions. Notably, there are various process mining tools available that exemplify this concept, such as Celonis, BehfaLab, and Appromore. It is important to note that while the listed modules serve as examples, it is worth acknowledging that different process mining tools may possess varying sets of modules. To delve further into these modules:

1. Dashboard Module: The Dashboard Module offers a comprehensive overview of process analytics and key performance indicators (KPIs).
2. Process Discovery Module: The Process Discovery Module takes event logs and translates them into visual representations of workflows.
3. Performance Mining Module: With the Performance Mining Module, one can assess process efficiency by identifying areas with high performance as well as those requiring improvement.
4. Conformance Checking Module: The Conformance Checking Module serves to ensure compliance of discovered process models with established standards.
5. Organizational Mining Module: Lastly, the Organizational Mining module plays a critical role in analyzing organizational structure and interactions—crucial elements for understanding the human aspects present within business processes.

It is essential to bear in mind that these listed modules provide merely an illustrative example; other process mining tools may incorporate different or additional modules tailored to their specific functionalities and purposes. Outputs of these modules are stored in databases for systematic data analysis and historical review.

- **Prompt Engineering for Enhanced AI Interaction**

A key aspect of the architecture is the integration of prompt engineering, structured as follows:

*1. Submodule Consideration:* The role of ChatGPT as an analyst and process mining expert is established, with an introduction to each module and its submodules (e.g., "act as a business process analyst").

*2. KPIs Description:* Descriptions of the Process Mining Engine's output KPIs are provided to ChatGPT (e.g., "Structural Analysis: Total cases: [value], Total activities: [value], Total variants: [value], Total cases with rework: [value]; Temporal analysis: First event date: [value], Last event date: [value]").

*3. Requested Task:* Specific requests for "Analytics" "Interpretations" and "Recommendations for Improvement" are made to ChatGPT, guiding the AI to provide targeted outputs.

- **Zero-shot and optimized prompt**

In the proposed architecture, zero-shot prompt engineering was initially explored to enhance the functionality of ChatGPT within the domain of process mining. Zero-shot prompt engineering enables ChatGPT to accurately respond to scenarios it has not encountered before, crucial for covering the broad spectrum of process mining challenges. This methodology is grounded in the concept of zero-shot learning, where AI models leverage their training to tackle new tasks without prior direct exposure (Xian et al., 2018). Although various prompt engineering techniques were considered to bolster ChatGPT's effectiveness, the focus shifted towards optimizing prompt engineering. This decision was influenced by the varied tasks and use cases in process mining, necessitating highly customized prompts. Upon evaluating different prompts, it became evident that optimized prompts delivered superior performance, making them more suited for the objectives of this research.

Optimized prompt engineering is central to the approach presented in this study, involving the careful creation and refinement of prompts to ensure that ChatGPT provides accurate and relevant responses for process mining tasks. This emphasis on optimization is based on comparative assessments that demonstrate its enhanced performance and better alignment with the intricate requirements of process mining. The inherent structure of both the Zero-Shot and Optimized prompt methodologies can be delineated as follows:

Zero-shot prompt Structure:

[**Role**] [**Task**] [**Process**] [**Organization**] [**Sector**] [**KPIs**] [**Objective**] [**Considerations**] [**Deliverables**] [**Analysis Guidelines**] [**Additional Instructions**] [**Module data**]

Optimized Prompt Structure:

[**Role**] [**Task**] [**Process**] [**Organization**] [**Analysis Focus**] [**Deep Dive**] [**Recommendations**] [**Additional Considerations**] [**Module data**]

Upon scrutiny of the above structures, it is evident that certain components are shared between them. Table 1 shows the explanation of each section of these prompts along with its example.

*Table 1: explanation of each section of prompts*

| Section | Explanation | Example |
|---|---|---|
| **Role** | The role section specifies the persona or professional capacity the AI should assume when responding to the prompt. This helps to frame the AI's perspective and expertise in the given context. | Role: Act as a seasoned financial analyst with expertise in evaluating corporate financial performance. |
| **Task** | The task section clearly defines the primary action or analysis the AI is expected to perform based on the given prompt. This guides the AI to focus on the specific task at hand and provide relevant outputs. | Task: Analyze the quarterly financial reports and identify key trends, ratios, and indicators that reflect the company's financial health. |
| **Process** | The process section indicates the specific business process or workflow that the AI should focus on while performing the task. This information helps to narrow down the scope of the analysis to a particular process within the organization. | Process: Examine the accounts receivable process, from invoice generation to payment collection, to identify potential bottlenecks and inefficiencies. |
| **Organization** | The organization section specifies the company, institution, or entity for which the AI is conducting the analysis. This context helps the AI to tailor its responses and recommendations to the specific needs and characteristics of the organization. | Organization: Perform the financial analysis for ABC Corporation, a multinational technology company headquartered in San Francisco, California. |
| **Sector** | The sector section provides information about the industry or market segment in which the organization operates. This context allows the AI to consider sector-specific factors, benchmarks, and best practices when performing the analysis and generating insights. | Sector: The analysis should take into account the unique challenges and opportunities within the automotive industry, such as shifting consumer preferences, regulatory pressures, and the rise of electric vehicles. |
| **KPIs** | The KPIs section lists the key performance indicators or metrics that the AI should consider while conducting the analysis. These KPIs serve as the foundation for the AI's evaluation and help to quantify the performance of the process or organization. | KPIs: The financial analysis should focus on the following key metrics: revenue growth, gross margin, operating margin, return on equity, debt-to-equity ratio, and cash flow from operations. |
| **Objective** | The objective section outlines the primary goals or purposes of the AI's analysis. This clarifies the desired outcomes and helps the AI to structure its response in a way that addresses these objectives effectively. | Objective: The main goals of the financial analysis are to assess the company's profitability, liquidity, and solvency, identify potential risk factors, and provide actionable recommendations for improving financial performance. |
| **Considerations** | The considerations section guides the AI to focus on specific aspects, challenges, or opportunities that are particularly relevant to the analysis. This ensures that the AI's insights and recommendations are comprehensive and address the most critical areas of concern. | Considerations: When analyzing the accounts receivable process, pay close attention to factors such as average collection period, bad debt expense, and the effectiveness of credit policies in mitigating default risk. |
| **Deliverables** | The deliverables section specifies the expected outputs or results of the AI's analysis. This helps to clarify the format, content, and level of detail required in the AI's response, ensuring that it meets the needs and expectations of the user. | Deliverables: The financial analysis should include a summary of key findings, visualizations of relevant trends and ratios, and a set of specific, actionable recommendations for optimizing financial performance. The insights should be presented in a clear, concise manner that can be easily understood by both financial and non-financial stakeholders. |
| **Analysis Guidelines** | The analysis guidelines section provides instructions on how the AI should approach the analysis and present its findings. This may include directives on the structure, style, and tone of the response, as well as any specific methodologies or frameworks to be employed. | Analysis Guidelines: When conducting the financial analysis, ensure that the insights are presented in a logical, structured manner. Use clear, concise language and avoid technical jargon where possible. Support your findings with relevant data points and benchmarks, and provide a balanced perspective that considers both short-term and long-term implications. |
| **Additional Instructions** | The additional instructions section includes any further directives or considerations that the AI should keep in mind while performing the analysis. These may encompass specific data handling requirements, assumptions to be made, or areas of focus that are not covered in the other sections of the prompt. | Additional Instructions: When analyzing the accounts receivable process, ensure that you consider the impact of seasonality on collections and adjust your recommendations accordingly. If any data anomalies or inconsistencies are identified, make note of these in your analysis and provide suggestions for further investigation or data validation. Focus on providing practical, implementable recommendations that align with industry best practices and the organization's strategic goals. |
| **Module data** | The data of five modules that are explained in the Process Mining Engine Modules section | comprehensive overview of process analytics and key performance indicators (KPIs) |

The integration of these advanced prompt engineering techniques— specifically, zero-shot and optimized prompting—into the proposed architecture represents a significant innovation in the application of AI in process mining. These techniques enhance ChatGPT's ability to understand and analyze complex process data, providing insightful and actionable outputs across a wide range of scenarios.

- **Integration with Conversational AI**

ChatGPT, connected via a robust API, receives structured outputs from the process mining modules. It means that due to the security of the raw data, they do not transfer to ChatGPT and the outputs of each module will be stored in a database and the mentioned database would integrate with ChatGPT through the APIs. It performs several functions:
- Analytics of Process: Analyzes data, providing insights on process performance and metrics.
- Interpretation of Process: Offers interpretations to identify trends and outliers.
- Recommendations for Improvement: Suggests strategies for process optimization.

- **User-Centric Design**

The architecture features a user-centric design, allowing interaction with ChatGPT through a conversational interface, making advanced data analytics accessible to a wide range of users. It facilitates the conversation of users with ChatGPT to ask more questions and request more insights about the process.

- **Staged Implementation and Empirical Validation**

The architecture was implemented in stages, with each module individually tested and validated. It was empirically evaluated using data from 17 companies using Process Mining tools, providing a robust dataset for system performance assessment and real-world feedback integration.

The proposed architecture in this study, which integrates Conversational AI, specifically ChatGPT, with process mining tools, offers several significant advantages. Primarily, it enhances user experience by transforming complex process mining data into more intuitive and accessible formats. This integration leads to improved readability and comprehensiveness of outputs, making it easier for users of varying expertise levels to understand and engage with the data. Additionally, the architecture facilitates the generation of actionable recommendations, enabling businesses to make informed decisions based on clear, concise, and relevant insights derived from their process data. By leveraging the advanced capabilities of Conversational AI, the architecture not only simplifies the interpretation of process mining results but also adds a layer of strategic intelligence to the decision-making process, thereby significantly contributing to the optimization and efficiency of business operations.

## 4. Results

This section presents the outcomes of the implementation of the proposed architecture, which integrates ChatGPT with process mining tools. To rigorously test this architecture, data was gathered from a Process Mining Tool, utilized by 17 distinct companies (Table 2). The evaluation process was methodically structured to ensure a comprehensive assessment of the architecture's effectiveness.

*Table 2: Comparative data of 17 companies*

| Sector | Economic Activity | Process |
|---|---|---|
| Public Sector | Municipality | Issuance Of Municipal License |
| | | Building Completion Approval |
| | | Unhindered, The District Has Taken Action |
| | | Unobstructed Area |
| | Public Administration | Order Submission |
| | Public Administration | Proceedings |
| | Public Administration | Legislative Activity |
| Service | Insurance | Payment Of Car Insurance Claims |
| | | Payment Of Medical Insurance Claims |
| | Post | Posting Goods |
| | Transportation | Airplanes Takeoffs and Landings |
| | Electricity Supply | Inspection Of Measuring Equipment |
| | Banking | Issuance Of Approval |
| | Technical Testing and Analysis | Technical Approval |
| Industrial | Manufacturing | Sales |
| | | Logistics |
| | | Maintenance |

A panel of experts, specializing in process mining and business analytics, was convened to evaluate the results. This panel played a pivotal role in scrutinizing the outputs generated by the integration of ChatGPT within the process mining framework. The outputs were systematically categorized for evaluation but are not explicitly detailed in this section to maintain focus on the methodology and the broader implications of the findings.

The evaluation conducted by the expert panel is crucial in contextualizing the results within the broader landscape of process mining and AI integration. This approach aligns with established practices in the field and ensures that the outcomes are assessed against a backdrop of existing literature and industry standards.

### 4.1. Process Mining Tool

BehfaLab is pioneering the use of AI-driven process mining in the Middle East. Launched in 2020, BehfaLab is the region's first platform to integrate artificial intelligence into process mining - offering a suite of capabilities to extract invaluable insights from system and human data.

The software contains six main modules empowering robust process analysis: A dashboard for visualization; Process Discovery to map workflows; Performance Mining to identify bottlenecks; Conformance Checking to find discrepancies; Organizational Mining connecting processes to people; and an AI module for predictive insights. Already recognized by ProcessMining.org and G2, BehfaLab enables organizations to optimize through technology-enhanced process intelligence. Its ability to illuminate workflows, quantify capacity, determine the root cause of variants, and learn baseline behaviors represents a major competitive edge for Middle Eastern enterprises navigating today's disruptive business landscape.

### 4.2. The Dataset

In evaluating the efficacy of the suggested architecture, which integrates ChatGPT with process mining tools, we conducted a thorough analysis of a dataset derived from a Process Mining Tool. This dataset comprises extensive data from 17 companies spanning diverse sectors, each presenting a distinct array of events and activities. The Public Sector includes data from 7 companies, with 428 activities and 473,945 events. The Service Sector, also with 7 companies, comprises 139 activities and 289,243 events. Lastly, the Industrial Sector, represented by 3 companies, includes 28 activities and 191,557 events. This distribution reflects the diverse nature of processes and activities across different industries, providing a rich context for evaluating the application of process mining and AI. Figure 3 shows the distribution of Activities and Events between sectors.

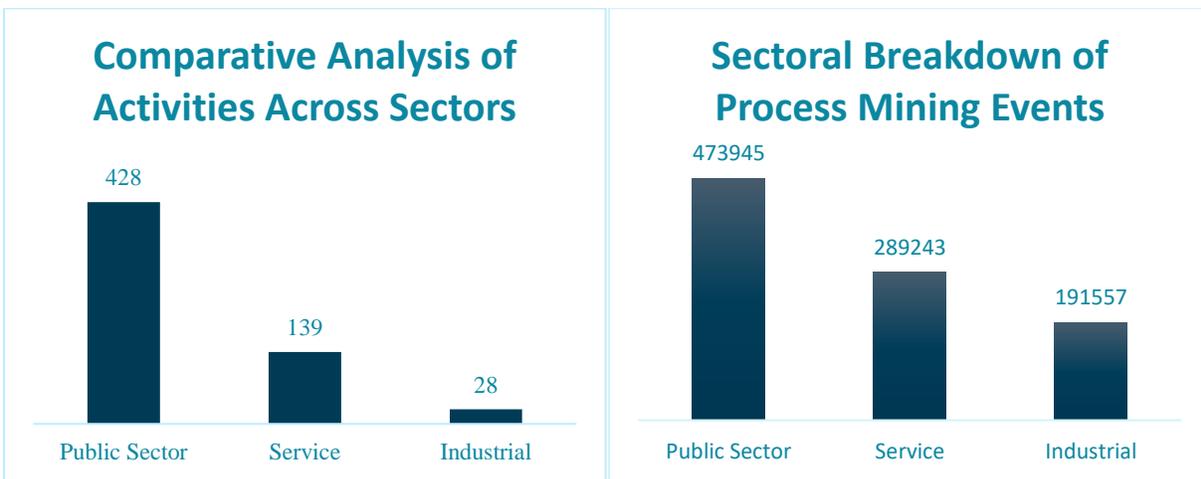

Figure 3: Distribution of Activities and Events between sectors

### 4.3. Evaluation Method

Expert panels are a cornerstone in evaluative and decision-making processes, particularly when specialized knowledge and diverse perspectives are required. Such panels are composed of individuals with significant expertise in relevant fields, their formation is often a response to complex, multifaceted issues where informed, expert opinions are invaluable. These panels engage in discussions, analyze data, and provide recommendations, thereby influencing key decisions and policies. Their effectiveness lies not only in the expertise of individual members but also in the collective wisdom that emerges from their collaborative deliberations. The diversity in composition — in terms of gender, age, professional background, and experience — enriches the panel's perspective, ensuring a holistic and nuanced understanding of the issues at hand. Expert panels operate in various modalities, from live discussions to remote collaborations, adapting to the logistical and contextual needs of the task.

In the context of this study, the evaluation methodology was significantly enhanced by the formation of a carefully selected panel of experts. Comprised of 10 professionals, including 7 males and 3 females from a spectrum of age groups and diverse work experiences as illustrated in Figure 4, the panel's role was crucial in assessing the integration of ChatGPT with process mining tools. Each member brought a unique perspective, grounded in their professional journey, contributing to a rich, multi-dimensional analysis of the technology integration. Their collective expertise, spanning various relevant domains, was instrumental in providing a comprehensive and informed evaluation, crucial for the robustness and validity of our study's findings.

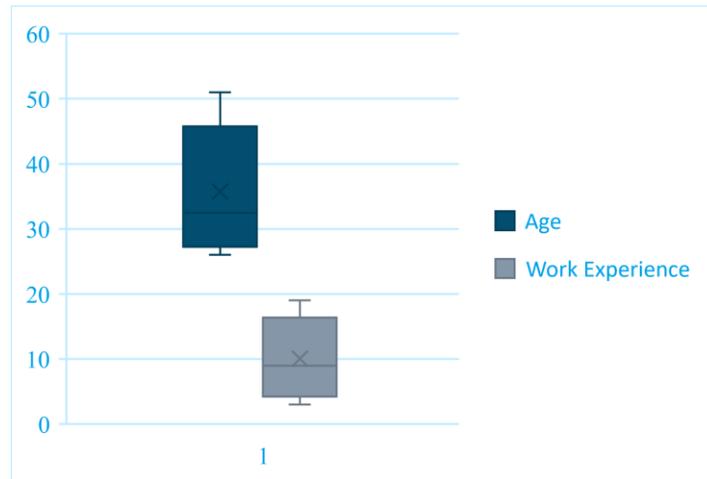

*Figure 4: Age and Work Experience Distribution of Expert Panel*

This panel was carefully selected to ensure a broad spectrum of expertise and perspectives in process mining and business analytics. The diversity in age and experience among the panel members provided a comprehensive and nuanced evaluation of ChatGPT's outputs. The use of expert panels in research methodology is a well-established practice, offering a systematic approach to gathering consensus and insights from seasoned professionals in a specific field (Evans, 1997; Hsu & Sandford, 2007).

Each expert applied their specialized knowledge to assess the relevance, accuracy, and applicability of ChatGPT's outputs, ensuring a rigorous and thorough evaluation process. The collective expertise of the panel not only lent credibility to the study's findings but also provided valuable insights into the practical implications of the research.

### 4.4. **Zero-Shot and optimized prompt results**

A comparative analysis delineates the enhancements in ChatGPT's performance through the application of zero-shot, few-shot, and optimized prompt engineering techniques within process mining. Initially, the investigation embraced a zero-shot approach, wherein ChatGPT engaged with new and unencountered scenarios, establishing a baseline of the model's capabilities.

The exploration subsequently progressed to few-shot learning. This phase, characterized by training the model with a constrained dataset, significantly improved ChatGPT's ability to generate precise and contextually relevant responses. However, the zenith of ChatGPT's performance enhancement was achieved during the optimized prompt engineering phase.

In this phase, prompts were rigorously tailored to align with the specific demands of process mining tasks, resulting in a marked improvement in ChatGPT's efficiency and accuracy. This advancement underscored the superiority of optimized prompt engineering in elevating the model's performance.

It is crucial to highlight that, despite the final implementation gravitating towards zero-shot and optimized prompts, a thorough evaluation of the few-shot learning outcomes was conducted. This assessment was instrumental in gauging the comparative effectiveness of each method, informing the decision to prioritize optimized prompts. Figure 5 visually encapsulates this progression, illustrating the stepwise enhancement in ChatGPT's effectiveness from the initial zero-shot approach to the refined and optimized prompt engineering, thereby evidencing the strategic selection process behind the prompt engineering approach to augment AI's utility in process mining.

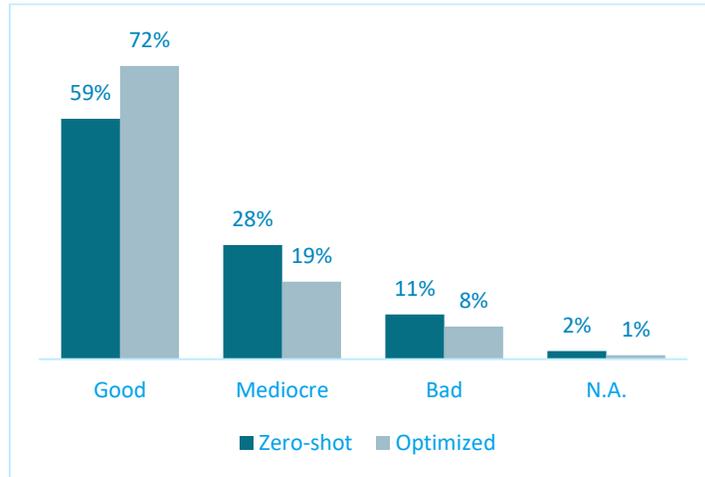

*Figure 5: Zero-shot vs Optimized Prompts*

The representation serves as a testament to the evolving capabilities of AI in the realm of process mining, illustrating how strategic advancements in prompt engineering can substantially elevate the analytical prowess of AI tools in complex data environments.

### 4.5. Results Distribution and Comparative Analysis

The evaluation of the proposed architecture integrating ChatGPT with process mining tools yielded results that are not only statistically significant but also offer deeper insights into the application of AI in process mining. The overall distribution of the results, as depicted in Figure 6, suggests a strong performance of the AI integration across various metrics:

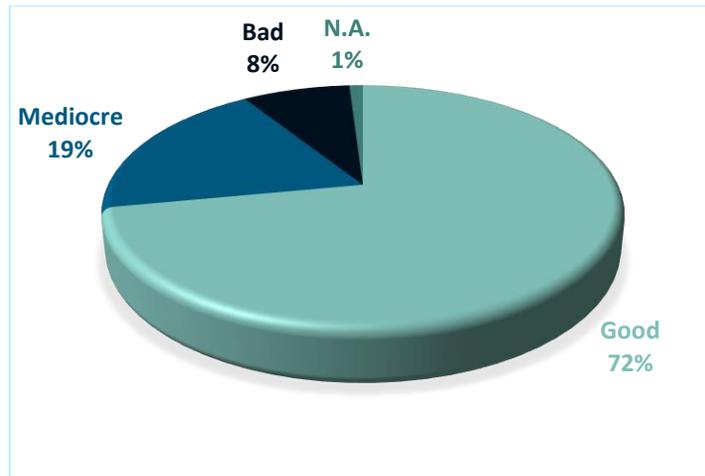

*Figure 6: Expert Panel Results*

Good: 72% of the results were "Good." This high percentage of "Good" results indicates that the integration of ChatGPT significantly enhances the quality of process mining outputs. It suggests that AI can effectively interpret complex data sets and provide actionable insights.
- Mediocre: 19% of the results were "Mediocre," suggesting areas for refinement but still showing a consistency of AI performance.

- Bad: 8% of the results were "Bad," which indicates that AI is well-integrated with process mining tools but it has works to be done.
- Not Available (N.A.): 1% of the results were "N.A.," indicating some problems in gathering answers from ChatGPT.

The sector-wise distribution of results, detailed in Figure 7, reveals interesting patterns:

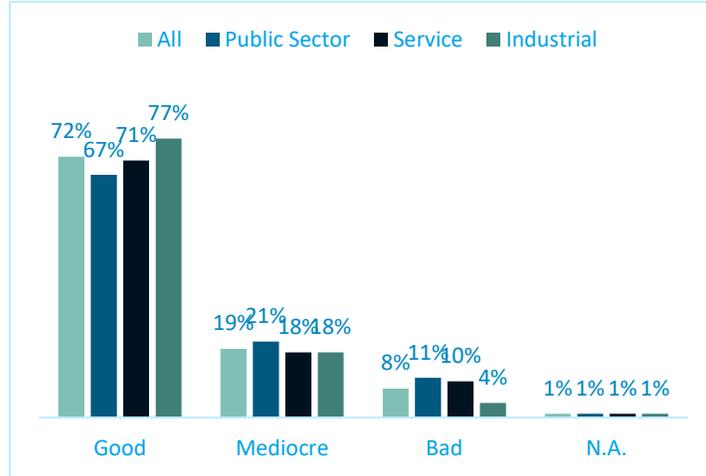

Figure 7: Expert Panel Results by Sectors

Public Sector (67% Good): The slightly lower percentage of "Good" results in the Public Sector may reflect the complexity and regulatory constraints inherent in public administration processes.

Service Sector (71% Good): The high percentage of "Good" results in the Service Sector indicates that AI is particularly adept at handling customer-oriented and service-based processes.

Industrial Sector (77% Good): The highest percentage of "Good" results in the Industrial Sector suggests that AI is extremely effective in environments with structured and repetitive processes, such as manufacturing.

The gender-wise distribution of results, shown in Figure 8, provides an additional layer of analysis:

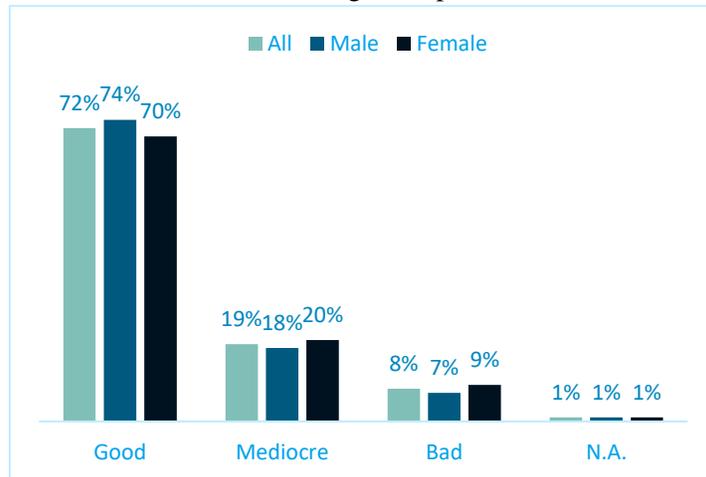

Figure 8: Expert Panel Results by Gender of Experts

Male Experts (74% Good): The marginally higher percentage of "Good" results among male experts could indicate varied perspectives or evaluation criteria based on gender.

Female Experts (70% Good): The close alignment of results between male and female experts underscores the consistency and reliability of the AI integration across different evaluators.

The comprehensive analysis of the results, considering overall distribution, sector-specific, and gender-specific breakdowns, not only validates the effectiveness of the AI integration but also provides critical insights into its applicability in different sectors and its perception among diverse expert groups. These insights are crucial for understanding the broader implications of AI in process mining and for guiding future enhancements and applications of AI in various business contexts.

### 4.6. Interpretation of Results and Contribution Validation

The predominance of "Good" results validates the effectiveness of integrating ChatGPT with process mining, marking a significant enhancement over traditional process mining tools and previous AI integrations (Van der Aalst et al., 2012). The results demonstrate the successful realization of the study's objective to democratize access to process mining analytics through conversational AI.

The presence of "Mediocre" and "Bad" results, while minimal, provides valuable insights for future improvements in prompt engineering and AI adaptability. The comparative analysis with prior works indicates a notable advancement in the field, aligning with the contributions outlined in this study.

In summary, the empirical evidence from the implementation across 17 companies using BehfaLab's Process Mining Tool underscores the efficacy of the proposed architecture. The integration of conversational AI with process mining tools, as evidenced by the high rate of "Good" results, represents a significant advancement in business process management, offering more intuitive, accessible, and effective tools for process analysis.

## 5. Discussion
### 5.1. Analyzing the Impact of ChatGPT on Process Mining

The integration of ChatGPT into process mining, as demonstrated in this study, marks a significant advancement in the field. The results from the 17 companies using BehfaLab's Process Mining Tool indicate a profound impact on the efficacy of conversational process mining tools. The high rate of "Good" outcomes (72%) underscores the effectiveness of ChatGPT in providing nuanced analytics, interpretations, and actionable recommendations.

### 5.2. Implications for Business Process Analysis Methodologies

The study's findings have significant implications for business process analysis methodologies. The ability of ChatGPT to transform complex process mining data into an intuitive conversational format represents a departure from traditional methods (Van der Aalst et al., 2012). This shift towards a more user-friendly approach in data analysis is crucial, especially in the context of conversational data analysis, where the integration of human-like interaction models can greatly enhance the analytical process. The advancements observed in this study suggest a future where conversational AI not only complements but also significantly enhances traditional process mining techniques.

## 6. Conclusion
### 6.1. Summarizing Key Contributions

This research has highlighted the transformative role of ChatGPT in the realm of conversational process mining. The integration of this advanced language model has led to substantial improvements in the depth and quality of analysis, interpretation, and recommendation generation. This goes beyond the capabilities of traditional process mining tools, offering a more nuanced and user-friendly approach to business process analysis (Van Der Aalst, 2016).

### 6.2. Advancing Business Process Analysis

The study contributes significantly to the field of business process analysis, particularly in the area of conversational data analysis. By integrating ChatGPT with process mining tools, the research has set a new standard in the application of conversational AI for business analytics. This represents a notable advancement from the methodologies discussed by (Dumas et al., 2018), highlighting the potential of AI to revolutionize traditional business process management practices.

### 6.3. Future Research Directions

As we look towards the future of process mining and AI integration, it's clear that the landscape is evolving rapidly. (Van der Aalst, 2023b) emphasizes the transformative potential of generative and predictive AI technologies in enhancing process management practices. Our findings align with this perspective, demonstrating the utility of ChatGPT in making process mining tools more accessible and effective. The case for integrating conversational AI into process mining is further validated by our empirical results, suggesting a promising avenue for future research in exploring advanced AI technologies and analytics to drive process optimization.

- **Acknowledgments**

We would like to express our sincere gratitude to Prof. Wil Van der Aalst for his invaluable feedback and guidance throughout the development of this manuscript. His thorough review and insightful comments, provided in two rounds, have significantly improved the quality and clarity of our work. We deeply appreciate his contributions and support.